# Sign Language Recognition based on YOLOv5 Algorithm for the Telugu Sign Language

**VIPUL REDDY.P[1], VISHNU VARDHAN REDDY.B[2], Dr.SUKRITI[3]**
[1]Electronics and Communication Engineering, Vellore Institute of Technology, Chennai
[2] Electronics and Communication Engineering, Vellore Institute of Technology, Chennai
[3] Senior Assistant Professor, SENSE, Vellore Institute of Technology, Chennai

**ABSTRACT** Sign language recognition (SLR) technology has enormous promise to improve communication and accessibility for the difficulty of hearing. This paper presents a novel approach for identifying gestures in Telugu Sign Language (TSL) using the YOLOv5 object identification framework. The main goal is to create an accurate and successful method for identifying TSL gestures so that the deaf community can use sign language to communicate with each other. To ensure precise localization of hand gestures and other visual clues, the research started by gathering an extensive dataset of TSL gestures that had been meticulously recorded. After that, a deep learning model was created that used the YOLOv5 architecture to recognize and classify gestures. This model benefited from the YOLOv5 architecture's high accuracy, speed, and capacity to handle complex sign language features. Utilizing transfer learning approaches, the YOLOv5 model was customized to TSL gestures. To attain the best outcomes, careful parameter and hyperparameter adjustment was carried out during training, giving priority to metrics like accuracy, precision, recall, and F1 score. With F1-score and mean Average Precision (mAP) ratings of 90.5% and 98.1%, respectively, the YOLOv5-medium model stands out for its outstanding performance metrics, demonstrating its efficacy in Telugu sign language identification tasks. Surprisingly, this model strikes an acceptable balance between computational complexity and training time to produce these amazing outcomes. Because it offers a convincing blend of accuracy and efficiency, the YOLOv5-medium model, trained for 200 epochs, emerges as the recommended choice for real-world deployment. The system's stability and generalizability across various TSL gestures and settings were evaluated through rigorous testing and validation, which yielded outstanding accuracy. This research lays the foundation for future advancements in accessible technology for linguistic communities by providing a cutting-edge application of deep learning and computer vision techniques to TSL gesture identification. It also offers insightful perspectives and novel approaches to the field of sign language recognition.

## I. INTRODUCTION

People who have trouble hearing mainly rely on sign language, which offers a structured communication system through hand gestures and facial expressions. Over 5% of people worldwide suffer from hearing impairment, and by 2050, that number is predicted to increase to 400 million, according to the World Health Organization (WHO)[21].Technology for sign language recognition (SLR) is essential to promoting inclusion and accessibility in communication for those with hearing loss. SLR systems let deaf people engage with the hearing world more successfully by translating sign language motions into text or voice. This helps to break down barriers to communication and encourages social inclusion. Telugu Sign Language is unique among the several sign languages in use globally since it is widely used in Telugu-speaking areas of India. In India, Telugu is the third most spoken language, with over 75 million native speakers. Telugu Sign Language is widely used, but not everyone is aware of it, which makes it difficult for deaf people to communicate effectively with others in the Telugu-speaking community. Therefore, improving communication accessibility and guaranteeing equitable participation for deaf people in Telugu-speaking cultures depends critically on the development of precise and effective methods for Telugu Sign Language recognition. When opposed to spoken language recognition, sign language identification poses distinct problems in spite of technological developments. Sign languages use elaborate hand gestures, face expressions, and body postures that display complicated visual and spatial grammar. Furthermore, the motions used in sign language varies greatly in various linguistic and cultural contexts, making standardization and recognition difficult. Computer vision and machine learning techniques are the mainstays of current approaches to sign language recognition. To recognize and decipher sign language motions, these methods usually combine handmade features, deep learning models, and annotated datasets. Although these techniques have showed potential, they frequently struggle to faithfully capture the subtleties and diversity of sign language expressions, particularly in real-world settings with varying illumination and distracting background noise.

## II. RELATED WORK

The need to improve communication accessibility for people with hearing impairments has prompted major developments in sign language recognition (SLR) technology in recent years. This paper delves deeply into foundational research and approaches in SLR, explaining how they apply to the creation of a Telugu Sign Language recognition system that makes use of the YOLOv5 object detection framework. The research referenced in [1] highlights the effectiveness of MobileNet architecture in Arabic sign language identification, outperforming earlier models in terms of accuracy and number of classes. This result highlights the potential utility of sophisticated neural network designs for Telugu Sign Language detection and offers a standard by which to compare performance with other models. It is shown in [2] that the MediaPipe model can track hand movements for Assamese Sign Language graphics with remarkable precision. This model is a strong contender for tracking Telugu Sign Language motions because to its scalability and resilience, especially in situations where proper detection depends on the precise location of hand movements. Explains a background removal method in [3] to improve the accuracy of character and digit recognition in sign language. The study's emphasis on enhancing recognition metrics through feature extraction and preprocessing aligns with the goals of Telugu Sign Language recognition, even though it focuses on several sign languages. Using comparable methods might improve Telugu Sign Language recognition models' accuracy.

Comparative studies like [4] offer important insights into the advantages and disadvantages of different SLR approaches, including as vision-based and data glove techniques. Telugu Sign Language recognition system design decisions are informed by an understanding of these trade-offs, which helps researchers choose the best method given the restrictions and goals of the project. [5] offers a sophisticated grasp of feature extraction and matching methods by exploring distance measures for sign feature matching. Effective feature matching algorithms are essential for correctly identifying Telugu Sign Language motions, especially when there are variances in the shape and movement of the hands. The viability of solutions based on Hidden Markov Models (HMM) for the recognition of Chinese sign language, as shown in [6], emphasizes the promise of probabilistic modeling techniques in SLR. The fundamental ideas of HMM-based recognition apply to Telugu Sign Language recognition as well, even if the study focuses on a different sign language. This offers a foundation for creating scalable and effective recognition models.

Advanced neural network architectures for dynamic sign language recognition are introduced in [7] and [8], including 3D Residual ConvNet and Bidirectional LSTM models. These models highlight the significance of temporal context in gesture identification tasks by effectively recognizing sign language gestures using spatiotemporal variables. A multi-scale perception strategy is proposed in [9] to achieve state-of-the-art recognition accuracy on large-scale continuous SLR datasets for weakly directed learning in video representation.

In [10], sign language recognition is turned into a video classification problem using a CNN and RNN model combination, with a 94% accuracy rate. While the LSTM model predicts signs and translates output to text, the CNN model gathers spatial data from brief movies. As shown in [11], the continuous sign language recognition using convolutional neural networks (CNNs) provides important insights into how to handle sequential data in SLR. It might be possible to recognize continuous sign language phrases and sentences by adapting this method to Telugu Sign Language. [12] offers a continuous sign language recognition system that uses deep neural networks and the ILSVRC-2014 dataset to achieve high accuracy. When developing continuous Telugu Sign Language recognition systems using the YOLOv5 framework, this paper is a useful resource. A CNN-LSTM model is used in [13] to demonstrate real-time recognition of the first six letters of Hindi varnamala, despite difficulties caused by background interference. This work clarifies the difficulties associated with real-time gesture identification and offers suggestions for reducing environmental disruptions to Telugu Sign Language recognition systems. [14] creates a web-based symbol decoding system with great accuracy by utilizing a refined version of the MobileNetV2 model and the Kaggle Dataset. This method demonstrates how web-based apps may be used to make Telugu Sign Language recognition systems available to a larger user population.

[15] uses background reduction and grayscale conversion to build a convolutional neural network model for the recognition of the 12 letters in Hindi varnamala, with

promising accuracy. This study emphasizes how crucial preprocessing methods are to boosting Telugu Sign Language recognition models' resilience. [16] achieves good results on benchmark datasets for sign language recognition using segmentation techniques and CNN-based classification. This approach can be modified to recognize Telugu Sign Language, making use of YOLOv5's object detection features to provide precise gesture localization. [17] and [18] investigate CNN models that achieve remarkable accuracy rates for the recognition of American Sign Language (ASL) and Indian Sign Language (ISL). Despite the fact that these research concentrate on various sign languages, their approaches and conclusions provide insightful information for the development of Telugu Sign Language recognition models.

Using Microsoft Kinect, a real-time sign recognition system in [19] records hand motions and uses SVM to achieve an average recognition accuracy. This study emphasizes the significance of reliable feature extraction techniques and shows that real-time Telugu Sign Language identification systems are feasible. In conclusion, [20] showcases a CNN-based model that was evaluated on a dataset of static signs and achieved remarkable training accuracy using several optimizers. With the YOLOv5 framework, this study offers insights into training methodologies and optimization techniques for creating Telugu Sign Language recognition models that are accurate and efficient.

These referenced studies collectively contribute to the foundational knowledge and methodologies in SLR, offering valuable insights and benchmarks for the development of a robust Telugu Sign Language recognition system using the YOLOv5 object detection framework. By leveraging advanced neural network architectures, preprocessing techniques, and feature extraction methods, researchers can design accurate and efficient Telugu Sign Language recognition models tailored to the unique characteristics of Telugu Sign Language.

### III. PROPOSED WORK

The major goal is to break down communication barriers and make it easier for people with disabilities to communicate. Hand gestures are a kind of mimed communication in sign language. Hearing and speech impaired people interact with one another through sign language, and numerous sign language detection technologies have been created to assist them in communicating with the general population. Our research looks at two such models. Our project involves identifying, translating, and converting hand gestures into alphabets.

#### A. Data Collection and Pre-Processing

We train and assess the model on our own recorded data set. The dataset includes 288 close-up, color photos. The photos are taken from 24 people under varied lighting circumstances, and image processing techniques are utilized to construct hand positions Figure 2. The dataset includes a total of 12 gestures. The Home (illu) , Love (prema) , Money (dabbulu) , No (kadhu) , One (okati) , Yes (avunu) , Fine (bagunna) , Family (kutumbam) , Pray (Namasthe) , Help (sahayam) , Why (enduku) , Where (ekkada) shown in Figure 1.

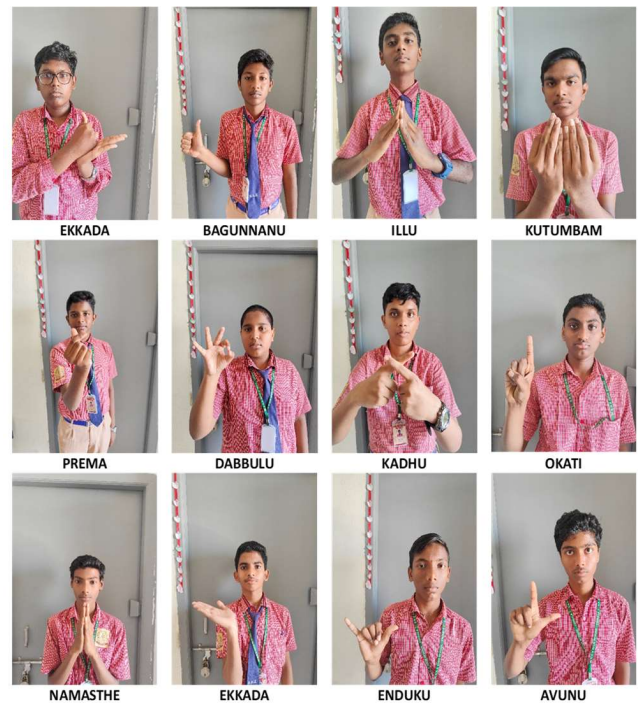

*Figure 1. Dataset*

The dataset includes photos of the gesture. To train data in YOLOv5, the data set must have labels and an annotating bounding box. The value of the annotation box coordinate should be normalized from 0 to 1. To create the bounding box for each image, labelimg software is utilized. This software allows you to easily annotate and create data labels in the preferred format. YOLO format for YOLO v5 and PyTorch.

The dataset is divided into three pieces. 80% of the data was for the training set, 20% for testing. The original dataset includes 288 photos. After partitioning the data set in this 80-20 ratio, the testing sets each have 58 photos, whereas the training set contains 230 images.

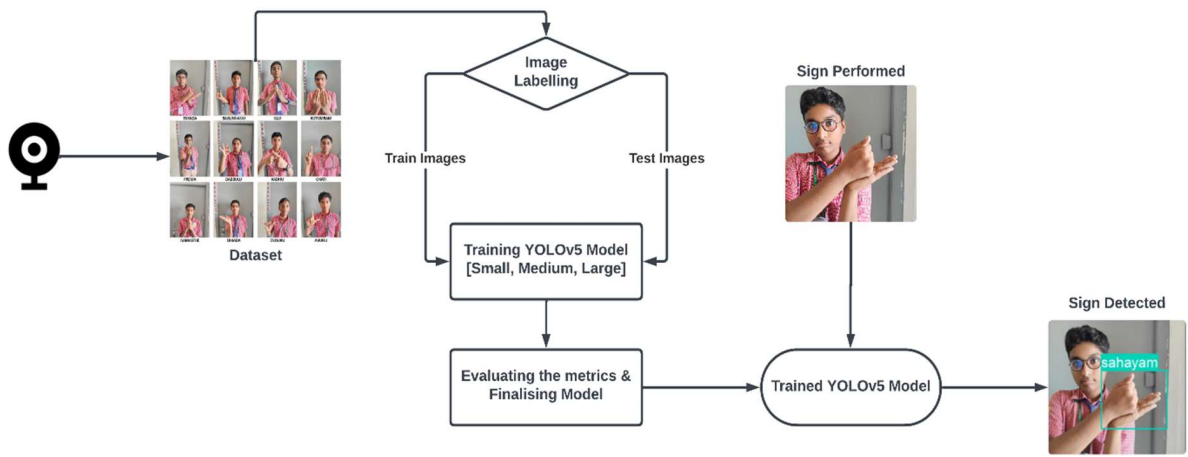

*Figure 2. Bock diagram*

### B. YOLOv5 Architecture

YOLOv5 design builds on the foundation of earlier versions of the YOLO (You Only Look Once) series, marking a significant leap in the field of object detection algorithms. It presents an object identification method that is both efficient and effective, achieving a careful equilibrium between simplicity, accuracy, and speed. Fundamentally, YOLOv5 predicts bounding boxes and class probabilities simultaneously by processing the full image in a single pass. Unlike conventional two-stage detection techniques, this simplified method does not require region proposal networks or post-processing stages.

YOLOv5's architecture shown as a visual representation below in Figure 3. consists of two primary parts: the detecting head and the backbone network. As the feature extractor, the backbone network processes the input image and extracts hierarchical information at various scales. Modern architectures like CSPDarknet53 or EfficientNet, which are selected for their efficacy and efficiency in capturing rich picture features, are usually the foundation of this network. Following their extraction by the backbone network, the features are sent to the detection head—a group of convolutional layers—for prediction-making. Using the retrieved features, the detection head predicts bounding box coordinates, objectness scores, and class probabilities. YOLOv5 can achieve real-time performance with high precision because to this simplified method.

The utilization of anchor boxes, which aid in enhancing localization accuracy by forecasting offsets in relation to preset bounding box forms and sizes, is a significant innovation of YOLOv5. Furthermore, YOLOv5 uses a multi-scale technique to efficiently capture objects of varied sizes, improving its resilience in various situations. YOLOv5 uses a combination of loss functions, such as terms for confidence loss, localization loss, and classification loss, to optimize its parameters during the training phase. The network will be able to precisely locate and identify items in the input image thanks to these loss functions.

YOLOv5 architecture offers a powerful blend of speed, precision, and simplicity, marking a substantial advancement in object recognition. It is well-suited for a variety of applications, including as autonomous driving, surveillance, and image understanding across several domains, thanks to its unique approaches and streamlined methodology.

YOLOv5 can be efficiently modified for tasks involving the recognition of sign language by considering every sign motion as a separate object class. Images of sign language gestures are used to train the architecture; each image has bounding box coordinates and associated class labels tagged on it. YOLOv5 gains the ability to precisely localize and categorize sign motions in real-time during training. YOLOv5's detection head is able to recognize whether sign gestures are present in the input image by predicting bounding box coordinates and objectness scores. Anchor boxes and multi-scale features let YOLOv5 capture the variation in size and appearance of sign motions. During the training phase, the network parameters are optimized to reduce the sum of the localization and classification losses. After training, the YOLOv5 model can be used for real-time inference on images or video streams, enabling prompt gesture identification and interpretation. High accuracy, real-time performance, and ease of deployment across several platforms are just a few benefits of this YOLOv5 modification for sign language detection, which helps to enhance accessibility and communication for the deaf community.

This experiment is carried out using YOLOv5, a deep learning architecture. YOLOv5 is lightweight, quick, and requires less computing resources than other current state-of-the-art architecture models, while maintaining accuracy comparable to current state-of-the-art detection models. It is substantially faster than previous YOLO variants. YOLOv5 extracts the feature map from the image using CSPNET as its backbone. It also utilizes the Path Aggregation Network (PANet) to improve information flow as shown in the Figure 3.

The graphic below depicts the architecture of YOLOv5. We use YOLOv5 for the following reasons:
1. Includes important features such as a cutting-edge activation function, hyperparameters, a data augmentation technique, and a user-friendly handbook.
2. Its lightweight architecture allows for computationally efficient training with limited resources.
3. The model's size is tiny and lightweight, making it suitable for usage with mobile devices.

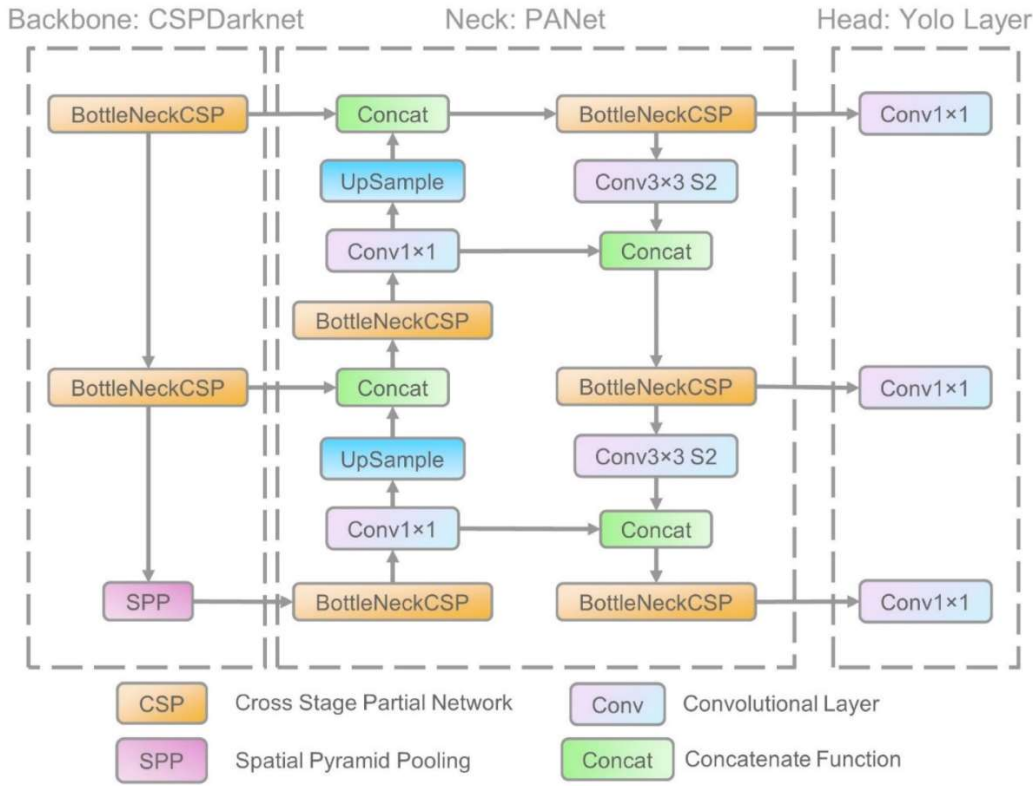

Figure 3. YOLOv5 acrhitecture [22]

$$Loss = \lambda_{coord}\sum_{i=0}^{S^2}\sum_{j=0}^{B}1_{ij}^{obj}[(x_i-\hat{x}_i)^2+(y_i-\hat{y}_i)^2] + \lambda_{coord}\sum_{i=0}^{S^2}\sum_{j=0}^{B}1_{ij}^{obj}[(\sqrt{w_i}-\sqrt{\widehat{w}_i})^2+(\sqrt{h_i}-\sqrt{\widehat{h}_i})^2]$$
$$+\sum_{i=0}^{S^2}\sum_{j=0}^{B}1_{ij}^{obj}(C_i-\hat{C}_i)^2 + \lambda_{noobj}\sum_{i=0}^{S^2}\sum_{j=0}^{B}1_{ij}^{noobj}(C_i-\hat{C}_i)^2 + \sum_{i=0}^{S^2}(p_i-\hat{p}_i)^2 \quad (1)$$

During training, this loss function strikes a balance between class probability estimate, objectless prediction, and localization accuracy. The YOLO object detection technique relies heavily on the YOLO (You Only Look Once) loss function, which is essential for teaching the model to recognize and categorize objects in a picture. The YOLO loss function is made up of multiple components that together indicate how well the model performs in tasks involving localization and classification. Localization loss, confidence loss, and classification loss are some of these concepts as shown in equation **(1)**. By penalizing erroneous object bounding box coordinate predictions, localization loss encourages the model to accurately localize items within the image. By punishing erroneous confidence scores for both object and background areas, confidence loss assesses the model's level of confidence in its predictions. Classification loss penalizes incorrect object category classifications and gauges how well the model can categorize items. The YOLO (You Only Look Once) model is an effective tool for a variety of computer vision applications because it learns to effectively detect and categorize objects in real-time by minimizing the combined loss across these terms during training using methods like gradient descent optimization.

Undoubtedly, the YOLOv5 loss function has a number of parameters that control the training process's optimization. These parameters regulate the relative relevance of localization and confidence losses, respectively, and contain the weight coefficients, $\lambda_{coord}$ and $\lambda_{noobj}$. The geographic granularity and number of predictions made by the model are determined by the grid size $S$ and the number of bounding boxes $B$ per grid cell. By determining whether a bounding box predictor is in charge of detecting an object within a grid cell, the indicator function, $1_{ij}^{obj}$, affects the contribution of localization and confidence losses. Additionally, the predicted bounding box coordinates ($x_i$, $y_i$, $w_i$, $h_i$), confidence scores $C_i$, and class probabilities, $p_i$ are compared against their corresponding ground truth values ($\hat{x}_i$, $\hat{y}_i$, $\hat{w}_i$, $\hat{h}_i$, $\hat{C}_i$, $\hat{p}_i$) to compute the loss. The YOLOv5 model optimizes for both localization accuracy and prediction confidence while learning to detect objects in images by varying these parameters and reducing the loss function using methods like gradient descent optimization.

## IV. THE PROPOSED MODEL

Several painstaking stages were needed in the approach for training the YOLO model using a custom sign language dataset. To guarantee access to the most recent iteration of the YOLO framework, the YOLOv5 repository was first cloned from the official ultralytics GitHub repository. After that, all dependencies needed to enable the training pipeline to run without a hitch were installed. After then, the dataset was carefully selected, and it included pictures of 12 different sign language courses. With the use of "Labelimg" software, these photos were located and labeled with the appropriate sign language movements. This ensured accurate annotation in the YOLO format, which defines bounding box coordinates and class labels. After the dataset was created and annotated, the data was split according to the 80-20 split convention, with 80% of the data going toward training and the remaining 20% going toward testing the model. As a result, reliable model generalization and performance evaluation on untested data were guaranteed. In addition, in order to expedite the training procedure, the number of classes was dynamically established by referencing the YAML file linked to the dataset, taking into account the particular sign language movements that were being studied.

A YAML file was carefully created, defining important parameters including the number of classes, depth and width multiples, and anchor box settings, in order to configure the YOLO model for training. By customizing these parameters to the specifics of the sign language recognition job, computational complexity was reduced and model performance was maximized. The YOLOv5 model architecture's depth is scaled by the "depth multiple" option. The model's depth is almost two-thirds of the default depth, as indicated by a value of 0.67. Complexity and capacity can be adjusted by varying the model's depth,'Width multiple' parameter scales the width of each layer in the model. With a value of 0.75, the model's layers are approximately three-quarters as wide as the default configuration. Adjusting layer width effects the model's capacity to capture nuanced characteristics. Anchor boxes are predefined bounding boxes applied by the model to anticipate object locations and sizes properly. There are three distinct sets of anchor boxes that are designed to fit distinct scales denoted by the factors P3/8, P4/16, and P5/32. Every set has pairs of width and height values that match the bounding box dimensions. The design of the YOLOv5 backbone, which is in charge of extracting features from input photos, is described in YOLOv5 Backbone. A sequence of convolutional layers make up the backbone, which is broken up by bottleneck blocks, also known as BottleneckCSP. These building blocks make it easier to grasp hierarchical features with progressively more complexity. The syntax '[from, number, module, args]' is used to specify the configuration of each tier in the backbone. This syntax specifies the number of layers, the kind of module (e.g., Focus, Conv), the source of the layer's input feature maps, and the arguments needed to instantiate the module. The YOLOv5 head's architecture, which produces bounding box and class probability predictions, is described in YOLOv5 Head. The convolutional, upsampling, and concatenation layers of the head are designed to refine and combine features that were extracted by the backbone. In the end, the head generates predictions at several scales (P3, P4, P5) by using the traits that were identified to forecast bounding boxes, confidence levels, and class probabilities for motions used in sign language.

Key parameters like the depth and breadth multiples were adjusted following trials with various YOLOv5 architecture configurations, including small, medium, and large models trained for varied numbers of epochs (100, 200, and 300). A depth multiple of 0.33 and a width multiple of 0.50 were selected for the YOLOv5-small model in order to produce a more condensed architecture appropriate for real-time applications. On the other hand, the YOLOv5-large model built a more intricate and potent network that could extract finer features from the input data by using a depth multiple of 1.0 and a width multiple of 1.0. But after careful analysis, the researchers found that the YOLOv5-medium model, trained for 200 epochs and with a depth multiple of 0.67 and a width multiple of 0.75, performed the best in terms of Mean Average Precision (mAP) score when taking training time and computational complexity into account. By choosing this model, the researchers made sure that the model could accurately recognize sign language gestures as shown in Figure 4 in a reasonable amount of time and that it would maximize both accessibility and computational efficiency for real-world deployment. Training with the carefully selected dataset and configured model architecture started after the model configuration was complete. YOLOv5-medium was chosen for training because of its ideal ratio of performance to computational efficiency. 200 epochs of training were done in order to balance the demands on computing power and model convergence. To assess the convergence and efficacy of the model, performance parameters like recall, accuracy, and loss were continuously tracked during training. The YOLO model was skillfully primed for the complex task of sign language recognition by carefully selecting the dataset, optimizing model parameters, and carrying out a rigorous training regimen. This cleared the path for accurate and effective communication accessibility for people who use sign language.

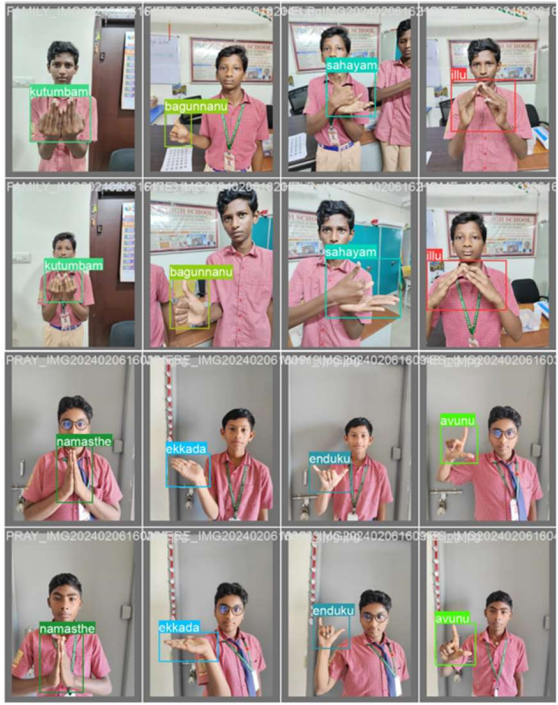

*Figure 4. Testing images*

## V. RESULTS AND DISCUSSION

We compare and analyze in detail the performance of YOLOv5 models, including small, medium, and large variations trained over 100, 200, and 300 epochs. The F1-confidence curve and mAP score, which are displayed in Table 1 and offer insights into the models' trade-offs between precision and recall across various confidence levels, are used to evaluate each model's efficacy.

The link between the F1 score, a harmonic mean of recall and accuracy, and different confidence thresholds used during inference is displayed on the F1-confidence curve. It is an important indicator for assessing how well the models perform overall in tasks involving object detection. Different patterns and trends show up when the F1-confidence curves for the YOLOv5-small, medium, and large models trained across various epochs are analyzed. The graphs show how the models perform when the detection threshold's confidence is changed. The F1-confidence curve for the YOLOv5-small model, trained over 100, 200, and 300 epochs, is displayed in Figure 5. It shows the model's precision-recall trade-offs at various confidence thresholds. Similarly, the F1-confidence curves that illustrate the performance of the YOLOv5-medium and large models, trained during the same epochs, are displayed in Figures 6 and 7, respectively.

*Table 1 mAP Score Summary*

| No of epochs | YOLOv5s | YOLOv5m | YOLOv5l |
|---|---|---|---|
| 100-epochs | 0.835 | 0.889 | 0.966 |
| 200-epochs | 0.947 | 0.981 | 0.97 |
| 300-epochs | 0.929 | 0.981 | 0.995 |

It is clear from a detailed analysis and comparison of the F1-confidence curves for each model that the YOLOv5-medium model, trained over 200 epochs, performs better than the other models. Notably, this model effectively minimizes false positives and false negatives while achieving the ideal balance between precision and recall across a range of confidence levels, demonstrating its effectiveness in reliably detecting objects. Moreover, temporal and computational complexity considerations support the selection of the YOLOv5-medium model at 200 epochs as the optimal model. This model is a sensible and effective option for real-world applications since it balances training time and computational resources while still giving outstanding performance.

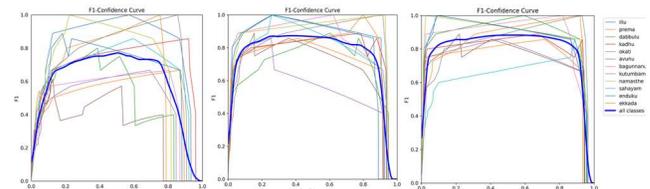

*Figure 5. YOLOv5s Epochs Comparison of f1-confidence curve*

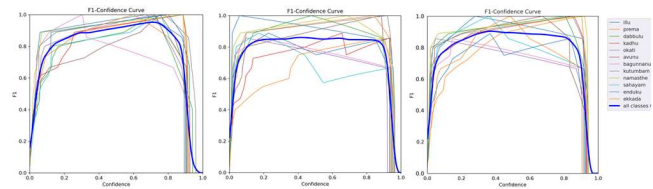

*Figure 6. YOLOv5m Epochs Comparison of f1-confidence curve*

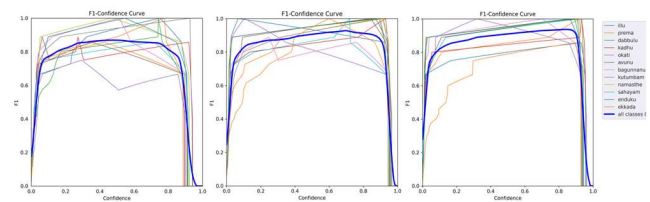

*Figure 7. YOLOv5l Epochs Comparison of f1-confidence curve*

Figure 8 presents a thorough examination of the performance indicators for the YOLOv5-medium model that was trained across 200 epochs. A visual depiction of several critical performance measures, such as mean Average Precision (mAP) score, object loss, class loss, precision, recall, and train and validation box loss, is given by the metrics graph.

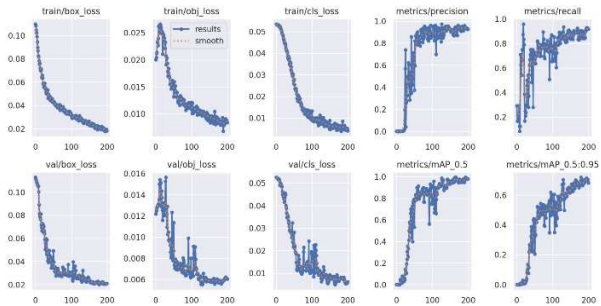

*Figure 8. Training metrics of YOLOv5m*

In the training and validation phases, the localization loss suffered is monitored using the train and validation box loss metrics, respectively. The model's accuracy in predicting the bounding box coordinates surrounding objects in the input images is assessed using this measure. Better localization accuracy is indicated by a reduced box loss. The confidence loss related to object presence predictions is measured by the train and validation object loss metrics. This measure evaluates how well the model can predict whether an object will be inside a bounding box. Better object detection performance is shown by lower object loss values. The classification loss that occurs during training and validation is assessed by the train and validation class loss metrics, respectively. This measure expresses how accurate the model's class predictions are. Better classification accuracy across several item categories is shown by a reduced class loss.

A performance statistic called precision measures the percentage of actual positive predictions among all the model's positive predictions. It gauges how well the model can prevent false positives, or situations in which it mispredicts the existence of an object. The percentage of true positive predictions among all actual positive cases in the dataset is measured by recall, which is sometimes referred to as sensitivity. In order to reduce false negatives, it assesses the model's capacity to identify every pertinent object in the input photos.

All item categories are combined into one complete statistic, the mean Average Precision (mAP) score, which combines recall and precision. To provide a single performance score, it first calculates the average precision for each class and averages them. Better overall performance in object detection is indicated by a higher mAP score. We can learn a great deal about the performance of the YOLOv5-medium model during various training and validation stages by examining the metrics graph. For the purpose of assessing the model's efficacy, resilience, and accuracy in object detection tasks, these measures are essential benchmarks.

*Table 2 Models Training Results for 100-epochs*

| Model | Precision(%) | Recall(%) | F1score(%) | mAP(%) |
|---|---|---|---|---|
| YOLOv5s | 74.2 | 74.8 | 74.5 | 83.5 |
| YOLOv5m | 92.2 | 78.2 | 84.6 | 88.9 |
| YOLOv5l | 90.3 | 92.3 | 91.3 | 96.6 |

$$Recall = \frac{TP}{TP + FN} \quad (2)$$

$$Precision = \frac{TP}{TP + FP} \quad (3)$$

$$F1 - Score = 2 \times \frac{Precision \times Recall}{Precision + Recall} \quad (4)$$

In order to use the trained YOLO model for sign language recognition, an image containing a sign language gesture must be sent to it. The convolutional neural network (CNN) layers that the YOLO model applies after receiving the input image are used to extract features and identify items in the image. In order to classify the observed objects—in this case, identifying the sign language gesture shown in the image—the model applies its learned parameters. As demonstrated in Figure 9, the YOLO (You Only Look Once) model visually highlights the sign's placement inside the image by drawing bounding boxes around it once it has been discovered. The YOLO model produces an output image that shows the original input image superimposed with bounding boxes around the movements that were identified in sign language. This clearly shows the model's recognition ability and sign class. For those with hearing problems, this approach makes it possible to detect sign language gestures in images quickly and accurately, improving communication accessibility.

*Table 3 Models Training Results for 200-epochs*

| Model | Precision(%) | Recall(%) | F1score(%) | mAP(%) |
|---|---|---|---|---|
| YOLOv5s | 93.2 | 88.5 | 90.8 | 94.7 |
| YOLOv5m | 90.9 | 90.2 | 90.5 | 98.1 |
| YOLOv5l | 92 | 89.7 | 90.8 | 97 |

*Table 5 COMPARISON OF RESEARCH PAPERS on Sign Language Recognition (SLR)*

| Developed SLR systems | Methodology | Dataset | Classes | Accuracy |
|---|---|---|---|---|
| N. F. Attia, A. E. Mostafa (2023) [23] | YOLO | MU HandImages ASL | 36 | 98 |
| G. Anilkumar, M. S. Fouzia (2022) [24] | YOLO & LSTM | 766 samples | 26 | 92.52 |
| B. Sanket, T. Kadam, K. korhale (2022) [25] | CNN & YOLO | 2150 samples | 26 | 88.4 |
| T. Dima and Md. E. Ahmed (2021) [26] | YOLO | MU HandImages ASL | 36 | 95 |
| R. Cui, H. Liu (2019) [11] | Bi-LSTM | rwth-phoenix and signum | 455 | 91.93 |
| Md. M. Rahman, M. S. Islam (2019) [12] | CNN | 154425 samples | 36 | 99 |
| J. Bora, S. Dehingia (2018) [2] | MediaPipe | 2094 samples | 9 | 99 |
| S. Das, Md. S. Imtiaz (2023) [3] | Background elimination and RF | 2080 samples | 46 | 91 |
| Y. Liao, P. Xiong (2019) [7] | BLSTM-3D | 25000 samples | 500 | 89.8 |
| Our Best result from Table 1 | YOLO v5 | 288 samples | 12 | 99.5 |
| Our Proposed result from Table 1 | YOLO v5 | 288 samples | 12 | 98.1 |

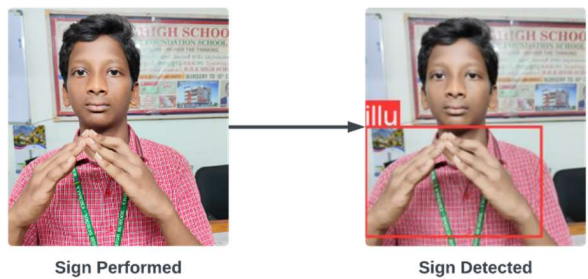

*Figure 9. Validation output diagram*

A comparative analysis of precision, recall, F1-score, and mean Average Precision (mAP) scores for the YOLOv5 models trained over 100, 200, and 300 epochs, correspondingly. is shown in Table 2, Table 3, and Table 5, provides an in-depth evaluation of the performance metrics for the different training durations. These tables provide insightful information on how the model's performance has changed over time, making it possible to monitor increases in object detection efficiency and accuracy as training epochs go up. These tables make it simple to compare the YOLOv5 models' performance across training phases and obtain a better grasp of how well they perform in sign language identification challenges.

*Table 5 Models Training Results for 300-epochs*

| Model   | Precision(%) | Recall(%) | F1score(%) | mAP(%) |
|---------|--------------|-----------|------------|--------|
| YOLOv5s | 97.4         | 83.1      | 89.7       | 92.9   |
| YOLOv5m | 95.2         | 93.9      | 94.5       | 98.1   |
| YOLOv5l | 95.4         | 89.1      | 92.1       | 99.5   |

## VI. CONCLUSION

Ultimately, the extensive analysis of YOLOv5 models over a range of epochs has confirmed that the YOLOv5-medium model, trained over 200 epochs, is the best option. This conclusion is the result of a thorough analysis that took computational efficiency and performance measures into account. The YOLOv5-medium model is particularly noteworthy for its exceptional performance metrics. Its efficacy in Telugu sign language recognition tasks is demonstrated by its 90.9%, 90.2%, 90.5%, and 98.1% precision, recall, F1-score, and mean Average Precision (mAP) scores, respectively. It's also remarkable that this model manages to balance training time and computing complexity and still provide such excellent outcomes. As a result, the 200 epoch-trained YOLOv5-medium model stands out as the best option for real-world deployment, providing an impressive blend of efficiency and accuracy in Telugu sign language recognition.

To sum up, this study has yielded significant findings regarding the use of the YOLOv5 model for sign language recognition. The most promising candidate has been determined to be the YOLOv5-medium model, which has been trained over 200 epochs. This model version balances computational complexity and training time, exhibiting greater accuracy in identifying and categorizing sign language gestures. These findings have important ramifications since they highlight how YOLOv5-based sign language identification systems can improve accessibility and inclusion for people with hearing loss and enable real-time communication support. Future research might investigate opportunities for improvement, such as larger datasets, hyperparameter fine-tuning, and alternative deep learning architectures, while also accepting the limits of this study, including dataset size and specific architectural decisions. All things considered, this research advances our understanding of assistive technology and computer vision, creating new opportunities for creative applications in accessibility and human-computer interaction.